%% file: main.tex
\documentclass[conference]{IEEEtran}
\usepackage{times}

\usepackage[numbers]{natbib}
\usepackage{multicol}
\usepackage[bookmarks=true]{hyperref}
\usepackage{xcolor}

\usepackage{graphicx}
\usepackage{float}
\usepackage{amssymb}
\usepackage{amsmath}
\usepackage{multirow}
\usepackage{tabularray}
\usepackage{caption}
\usepackage{subcaption}
\usepackage{booktabs}
\usepackage{colortbl}
\usepackage{hyperref}
\usepackage{cuted}
\usepackage{capt-of} 
\usepackage{booktabs}
\usepackage{pifont}
\usepackage{makecell}

\newcommand{\cmark}{\ding{51}}
\newcommand{\xmark}{\ding{55}}

\newcommand\thiswork{StereoVLA}
\newcommand\featurefusion{GeoSem Vision Encoder}
\newcommand\depthestimation{interaction-region depth estimation}
\newcommand\cameraestimation{camera parameter estimation}
\newcommand\benchmark{LIBERO-MV-R}

\begin{document}
\IEEEoverridecommandlockouts

\title{StereoVLA: Enhancing Vision-Language-Action Models with Stereo Vision%
\thanks{\hspace*{-1.4em}$^{*}$ Equal contribution with the order determined by rolling dice. $^{\dagger}$ denotes corresponding authors.\protect\\
Correspondence to \href{mailto:zhangzz@galbot.com}{zhangzz@galbot.com}, \href{mailto:hewang@pku.edu.cn}{hewang@pku.edu.cn}.}}

\author{Shengliang Deng$^{1,4*}$  Mi Yan$^{1,2*}$  Yixin Zheng$^{1,3*}$  Jiayi Su$^{1,5}$ Wenhao Zhang$^{1,2}$ \\ Yitao Zeng$^{1,6}$ Xiaoguang Zhao$^{4}$  Heming Cui$^{3}$  Zhizheng Zhang$^{1\dagger}$  He Wang$^{1,2\dagger}$\\
{\small $^{1}$Galbot $^{2}$CFCS, School of CS, Peking University $^{3}$Institute of Automation, Chinese Academy of Sciences} \\
{\small $^{4}$The University of Hong Kong $^{5}$Xiamen University Malaysia $^{6}$Southern University of Science and Technology}%
}



%

\maketitle

\input{fig/teaser_image}
\input{sec/0_abs}

\IEEEpeerreviewmaketitle

\input{sec/1_intro}
\input{sec/2_related}
\input{sec/3_method}
\input{sec/4_exp}
\input{sec/5_limitation}
\input{sec/6_conclusion}

\section*{Acknowledgments}
This work was partially supported by New Generation Artificial Intelligence-National Science and Technology Major Project (No. 2025ZD0122905), and National Key R\&D Program of China (2022ZD0160201). Additionally, we thank Yitao Zeng for assistance with the experiments.


\bibliographystyle{plainnat}
\bibliography{references}

\clearpage
\input{sec/7_supp}

\end{document}

%% file: fig/teaser_image.tex
\begin{strip}
    \centering
    \vspace{-40pt}
    \includegraphics[width=\textwidth]{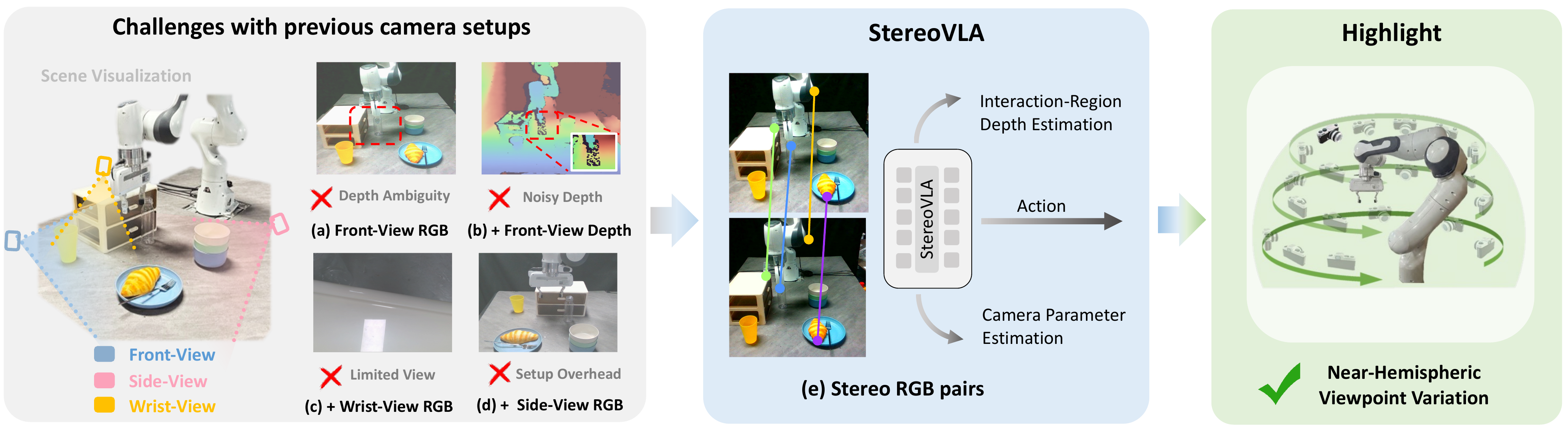}
    \captionof{figure}{
        \textbf{Left:} To enhance spatial perception, prior vision-language-action models typically supplement single-view RGB with sensor depth, wrist-view RGB, or side-view RGB. These setups face inherent challenges: (a) single-view RGB suffers from depth ambiguity; (b) depth sensors yield noisy estimates for transparent objects; (c) wrist-view RGB captures a limited view and the additional hardware increases collision risk; (d) multi-camera settings incur additional deployment overhead. \textbf{Middle:} (e) Stereo RGB pairs provide robust spatial cues for comprehensive scene understanding, mitigate view limit and collision risk, while enabling a simpler setup with a single stereo camera. Leveraging these distinctive strengths, we present StereoVLA, a novel VLA framework that takes in rich spatial and geometric cues from stereo RGB pairs, coupled with a dual-objective co-training strategy. \textbf{Right:} StereoVLA demonstrates robustness to near-hemispheric camera perspectives.
    }
    \label{fig:teaser}
    \vspace{-10pt}
\end{strip}

%% file: sec/0_abs.tex
\begin{abstract}

While Vision-Language-Action (VLA) models excel in generalist manipulation, they often lack fine-grained spatial awareness and show limited viewpoint robustness.
This limitation largely stems from the reliance on pretrained RGB encoders, which lack explicit geometric cues and prioritize semantic alignment over geometric representation. 
We argue that effective visual representations for VLA models must jointly encode both semantic and geometric information. 
In this paper, we introduce StereoVLA, the first VLA model to incorporate rich geometric cues from large-scale synthetic stereo data. 
StereoVLA employs a Geometric-and-Semantic (GeoSem) vision encoder that extracts geometric cues from subtle stereo-view disparities for precise spatial perception, while simultaneously capturing semantic features from pixel observations to support language-conditioned manipulation. 
Additionally, we introduce two synergistic co-training objectives: Interaction-Region Depth Estimation for precise spatial reasoning, and Camera Parameter Estimation to implicitly align perception and action coordinate systems. 
Compared with baselines that employ various input modalities, StereoVLA achieves a 33.4\% absolute gain in success rate in real-world experiments and demonstrates robustness to near-hemispheric camera perspectives.
Project page: \href{https://shengliangd.github.io/StereoVLA-Webpage/}{shengliangd.github.io/StereoVLA-Webpage/}.




\end{abstract}

%% file: sec/1_intro.tex
\section{Introduction}

Leveraging powerful pretrained Vision-Language Models (VLMs)~\cite{paligemma, qwen2.5-vl}, VLA models can effectively acquire diverse skills from demonstrations while exhibiting robust semantic reasoning. Nonetheless, pioneering VLA models (e.g., OpenVLA~\cite{openvla}) are confined to single-view RGB inputs, limiting their fine-grained spatial awareness and viewpoint robustness. Consequently, providing VLA models with rich geometric cues remains a key challenge.

Previous works explored three types of additional sensors: wrist-mounted cameras~\cite{pi0.5,groot,rdt}, depth sensors~\cite{rvt,3d-vla}, and extra third-person cameras~\cite{graspvla}. However, as shown in Figure~\ref{fig:teaser}, wrist cameras provide limited views that are easily occluded, and the wrist-mounted hardware increases collision risk with the environment. 
The reliability of depth sensors is often compromised by surface properties (e.g., transparent, specular, or light-absorbing materials) and external lighting conditions (e.g., sunlight saturation).
Extra third-person cameras offer complementary information, but they complicate hardware for mobile robots and humanoids, and the increased view diversity hinders viewpoint robustness~\cite{decomposing}.
These settings do not explicitly provide stereoscopic perception for embodied manipulation, nor learning methods that are well aligned with such perceptual cues. 


Beyond the limitations imposed by camera configurations, we further observe that most existing VLA models rely on pretrained RGB vision encoders, which lack effective representations of geometric information. However, effective vision representations should jointly encode semantic and geometric information to ensure that robots can not only understand instructions, but also interact precisely with objects of diverse shapes, positions, and other geometric properties. Given these limitations, we explore enhancing VLA model with stereo vision to enable more precise robotic manipulation. While prior work incorporates only limited stereo data in a straightforward manner~\cite{droid}, and recent vision research has developed strong stereo-based depth foundation models~\cite{foundationstereo}, how to learn and integrate effective stereo representations into VLA models remains an important yet underexplored research problem.

In this paper, we introduce StereoVLA, the first VLA model to incorporate rich geometric cues from large-scale synthetic stereo data. We experimentally find that directly applying existing RGB encoders to stereo inputs based on current state-of-the-art VLA models, such as~\cite{pi0.5}, is not sufficiently effective for learning embodied manipulation. This stems from two primary reasons. First, existing VLA models rely on pretrained RGB encoders, which lack explicit geometric cues and prioritize semantic alignment over geometric representation. Second, collecting large-scale robot data with stereo inputs is labor-intensive.

To address afore-discussed issues, we design a Geometric-and-Semantic (GeoSem) vision encoder to learn informative visual representations from a large-scale simulated data. The \featurefusion{} extracts geometric cues from subtle stereo-view disparities for precise spatial perception, while simultaneously capturing semantic features from pixel observations to support language-conditioned manipulation. Specifically, the \featurefusion{} employs the learned geometry-centric features from FoundationStereo~\cite{foundationstereo}, a state-of-the-art depth estimation foundation model pretrained on millions of stereo pairs.
It further integrate and spatially align semantically rich tokens from PrismaticVLM~\cite{prismatic} with these geometry-centric features to generate hybrid tokens that effectively combine geometry precision with semantics richness.

Besides, to tackle the challenge of limited stereo data, we draw inspiration from recent works~\cite{graspvla, robotwin, interndata, geniesim3} to generate 5 million synthetic stereo trajectories. These works demonstrate that the synergy between modern high-fidelity simulators and powerful vision-language foundation models can effectively bridge the sim-to-real gap, enabling seamless zero-shot sim-to-real transfer. To mitigate visual and physical gaps, we utilize extensive domain randomization and adopt a quasi-static assumption~\cite{mason2001mechanics}. We also observe that while modern rendering excels in RGB photorealism, it struggles to replicate sensor-specific depth artifacts, positioning stereo as a more robust modality for geometric sim-to-real transfer.


Furthermore, to improve viewpoint robustness, we rethink the perception-to-action mapping by decoupling it into camera-frame spatial perception and robot-frame action generation, enhanced respectively by two co-training tasks. 
First, \textbf{\depthestimation{}} guides the model to predict depth within interaction regions, rather than uninformative backgrounds, to learn task-relevant spatial details.
Second, \textbf{\cameraestimation{}} enables the model to infer camera parameters, learning the mapping between camera and robot frames implicitly.

In summary, our contributions are as follows: a) we present \thiswork{}, a geometry-aware VLA system that exploits rich geometric cues from stereo vision by integrating stereo sensing, stereo-aware representation learning, auxiliary geometric supervision, and large-scale synthetic-data pre-training for robust and precise manipulation, b) we introduce \featurefusion{} to produce visual tokens that carry dense semantic and geometric features from stereo inputs, c) we propose co-training with \depthestimation{} and \cameraestimation{} tasks to facilitate perception-to-action mapping with enhanced viewpoint robustness, d) we show that our approach outperforms existing VLAs by a large margin in diverse tasks under stereo settings, and is robust to near-hemispheric camera pose variations.
Evaluation shows that our approach significantly outperforms baselines under stereo configuration, achieving a 33.4\% higher success rate on general tasks and notable improvements in precision-demanding scenarios, and demonstrates robustness to near-hemispheric camera pose variations. Ablation studies validate the effectiveness of the proposed components.

%% file: sec/2_related.tex
\section{Related Work}

\subsection{Vision-Language-Action Models}

Recent advances in vision-language models~\cite{palmE,prismatic,eagle,qwen2-vl,qwen2.5-vl,qwen3-vl,florence2} and large-scale robot datasets~\cite{oxe,agibot_world,droid,rh20t,robomind} have accelerated Vision-Language-Action (VLA) models~\cite{RT2,openvla,pi_0,pi_fast,pi0.5,dexvla,hybridvla,cogact,graspvla,internvla-M1,tinyvla} for generalizable multimodal reasoning and control. Early methods~\cite{RT2,openvla} discretized actions for autoregressive prediction, whereas subsequent work~\cite{pi_0,pi_fast,pi0.5,dexvla,hybridvla,cogact,graspvla,internvla-M1} increasingly adopts diffusion or flow-based policies for smooth trajectories.

VLA camera setups have also evolved from single RGB views~\cite{RT2,openvla} to wrist-mounted monocular cameras~\cite{pi_0,pi_fast,dexvla,hybridvla,internvla-M1}, side-facing views~\cite{graspvla,cogact}, and multi-view layouts such as the four-corner configuration in $\pi_{0.5}$~\cite{pi0.5}. Despite this shift toward multi-view sensing, prior VLA work has not leveraged stereo cameras which can provide explicit geometric cues via binocular disparity.

Synthetic data is becoming a key ingredient for scalable VLA training, alleviating real-world data collection costs. With modern simulators and automated generation pipelines~\cite{genmanip,robocasa,robotwin,geniesim3,interndata}, recent efforts~\cite{graspvla,simandrealcotraining} show that combining synthetic trajectories with strong vision-language backbones improves real-world manipulation. Complementary lines of work explore real-to-sim-to-real augmentation to increase data diversity and robustness~\cite{xsim,real2edit2real,demogen,mimicgen,robosplat,real2render2real,fieldgen,exogs}.

\subsection{Robot Learning with Geometry Cues}

To address the limits of monocular RGB perception, a growing line of work~\cite{3d-dp,idp3,pointbridge} incorporates explicit geometry cues such as depth, point clouds, and reconstructed 3D structure. These geometry-aware inputs provide stronger spatial priors and improve robustness and generalization. For example, DP3~\cite{3d-dp} shows point-cloud observations can be more robust to lighting and object variations, and iDP3~\cite{idp3} further improves robustness by using a robot-centered egocentric frame that better handles environmental changes and supports both egocentric and third-person viewpoints.

\input{fig/pipeline_image}

Motivated by these advances, recent VLA methods have started to combine geometry with language and action to mitigate the spatial reasoning limitations of purely 2D models~\cite{3d-vla,spatialvla,pointvla,bridgevla,fp3,lift3d,evo,gp3,geovla,depthvla,qdepthvla,vlaser,peafowl,pragmaticvla,4dvla}. Representative examples include PointVLA~\cite{pointvla}, which injects depth cues from RGB-D or monocular depth estimation, and BridgeVLA~\cite{bridgevla}, which projects 3D observations into multi-view 2D feature maps for action prediction. However, existing methods still leave open how to best fuse explicit geometry with the rich 2D representations of pretrained vision foundation models.


\subsection{Stereo Perception and Policies for Robot Learning}

Deep stereo matching has progressed from accurate but memory-intensive cost-volume 3D CNNs~\cite{chen2023learning,cfnet,pcw,aanet,ral_stereo_depth_cost_volume,ral_stereo_depth_cost_volume_2} to iterative refinement~\cite{raft,iterative,li2022practical,jing2023uncertainty,gong2024learning,ral_stereo_depth_raft_style} avoiding explicit 4D volumes; recent work such as FoundationStereo~\cite{foundationstereo} further improves generalization via large-scale pretraining.

Stereo has also been used for manipulation policies, including stereo visual servoing~\cite{stereo-visual-servoing}, distillation-based visuomotor control (DextrAH-RGB)~\cite{dextrah-rgb}, and synthetic-data-driven perception for robust grasping (SimNet)~\cite{simnet}.

Different from these typically small-scale stereo-policy frameworks, we scale stereo perception to large VLA models, combining model capacity with vision-language grounding for stronger spatial awareness and cross-task generalization.


%% file: fig/pipeline_image.tex
\begin{figure*}[htbp]
    \centering
    \includegraphics[width=\linewidth,trim={0 0.8cm 0 0},clip]{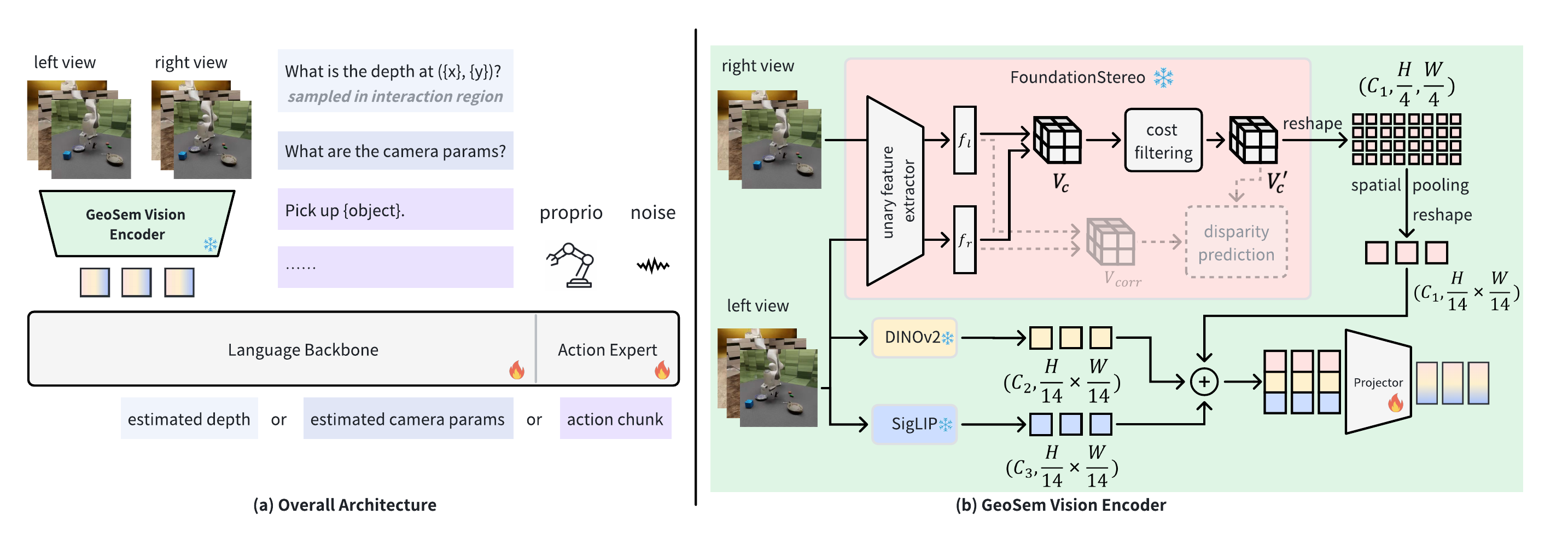}
    \caption{(a) In \thiswork{}, a stereo image pair is encoded by the \featurefusion{} module to generate visual tokens with geometric precision and semantic richness. Together with language tokens, they are processed by a large language model backbone (InternLM-1.8B). An action expert predicts delta end-effector poses, while auxiliary depth and camera parameter estimation tasks
    further enhance spatial reasoning during training. (b) The \featurefusion{} module extracts geometric features with FoundationStereo (bypassing disparity prediction components for efficiency) and semantic-rich features with SigLIP and DINOv2, then fuses them into a unified visual representation with an MLP projector.}
    \label{fig:model-arch}
\end{figure*}

%% file: sec/3_method.tex
\section{Method}

The overall architecture of \thiswork{} is illustrated in Figure~\ref{fig:model-arch}(a).
The \textbf{\featurefusion{}} module leverages state-of-the-art vision foundation models to encode the stereo images into visual tokens that capture both semantic and geometric information. These visual tokens, together with language tokens, are passed to a pre-trained large language model, InternLM-1.8B~\cite{internlm}, for vision-language joint processing. Leveraging the intermediate vision-language features (i.e., the key-value cache), a 300M-parameter action expert predicts action chunks via flow-matching~\cite{flowmatching, pi_0} with a delta end-effector pose representation. To improve viewpoint robustness, we introduce two synergistic tasks, \textbf{\depthestimation{}} and \textbf{\cameraestimation{}}, co-trained with action prediction to enhance spatial reasoning while preserving computational efficiency.

\subsection{\featurefusion{}}
As shown in Figure~\ref{fig:model-arch}(b), given a stereo image pair $I_l, I_r \in \mathbb{R}^{H \times W \times 3}$, we carefully extract geometric features from FoundationStereo \cite{foundationstereo} from both views and semantic-rich features from the monocular left view $I_l$. These features are then fused into a sequence of visual tokens that jointly encode both geometric and semantic information.

\textbf{Geometric Feature Extraction.}
FoundationStereo is tailored for depth estimation with several specialized modules, making feature adaptation to our task non-trivial. We therefore first outline the key components of FoundationStereo and then explain our feature choices.

For stereo image pairs $I_l, I_r$, FoundationStereo first computes  monocular features $f_l, f_r$ for each image with a unary feature extractor. These features are then processed via two parallel paths. In the first path, $f_l, f_r$ are concatenated to form a 4D cost volume $V_c \in \mathbb{R}^{C \times \frac{D}{4} \times \frac{H}{4} \times \frac{W}{4}}$, where $H$ and $W$ denote the image height and width, $D$ is the maximum disparity range considered, and $C$ is the feature dimension. To model long-range correlations, the cost volume undergoes an attention-based hybrid cost filtering module, yielding a filtered cost volume $V_c'$ with the same dimensions. In the second path, the dot product of the left and right unary features $f_l, f_r$ are computed to generate a correlation volume $V_{corr} \in \mathbb{R}^{\frac{W}{4} \times \frac{H}{4} \times \frac{W}{4}}$. Using the filtered cost volume $V_c'$ and correlation volume $V_{corr}$, FoundationStereo iteratively refines disparity estimates, resulting in a final disparity map.

When selecting features, we consider two criteria: (1) they should be dense feature volumes and capture rich geometric information, and (2) additional computation for depth estimation should be minimized. 

Guided by these criteria, we choose the filtered cost volume $V_c'$ as our geometric feature source. Although the iterative refinement in FoundationStereo further improves disparity estimation, it introduces considerable computational overhead. Similarly, the correlation volume $V_{corr}$ consists of matching scores rather than dense features, so we discard the second computation path. From the first path, we choose $V_c'$ over the raw cost volume $V_c$ because it incorporates long-range correlations via the hybrid cost filtering module, providing dense geometric features crucial for manipulation.

\textbf{Semantic Feature Extraction.}
As FoundationStereo is pretrained for depth estimation, it lacks the semantic and appearance information for effective vision-language grounding.
To overcome this, we extract these features using SigLIP and DINOv2 following PrismaticVLM~\cite{prismatic}.
SigLIP, trained with a vision-language contrastive objective, effectively captures high-level semantics.
DINOv2, trained with self-supervised discriminative objectives that align features across different image crops, excels at capturing visual details~\cite{dinov2,prismatic}.
Due to the high redundancy of the semantic information between left and right views, we apply SigLIP and DINOv2 exclusively on the left view for computational efficiency.

\textbf{Feature Fusion.}
The spatial resolution of FoundationStereo features differ with those of SigLIP and DINOv2. We therefore spatially pool the FoundationStereo features to match the 14-pixel stride of the other two. The final hybrid representation is obtained by concatenating the feature maps along the channel dimension. We avoid token sequence concatenation, as it increases the number of tokens and computational overhead during both training and inference.

\begin{figure}[t]
    \centering
    \includegraphics[width=0.80\linewidth]{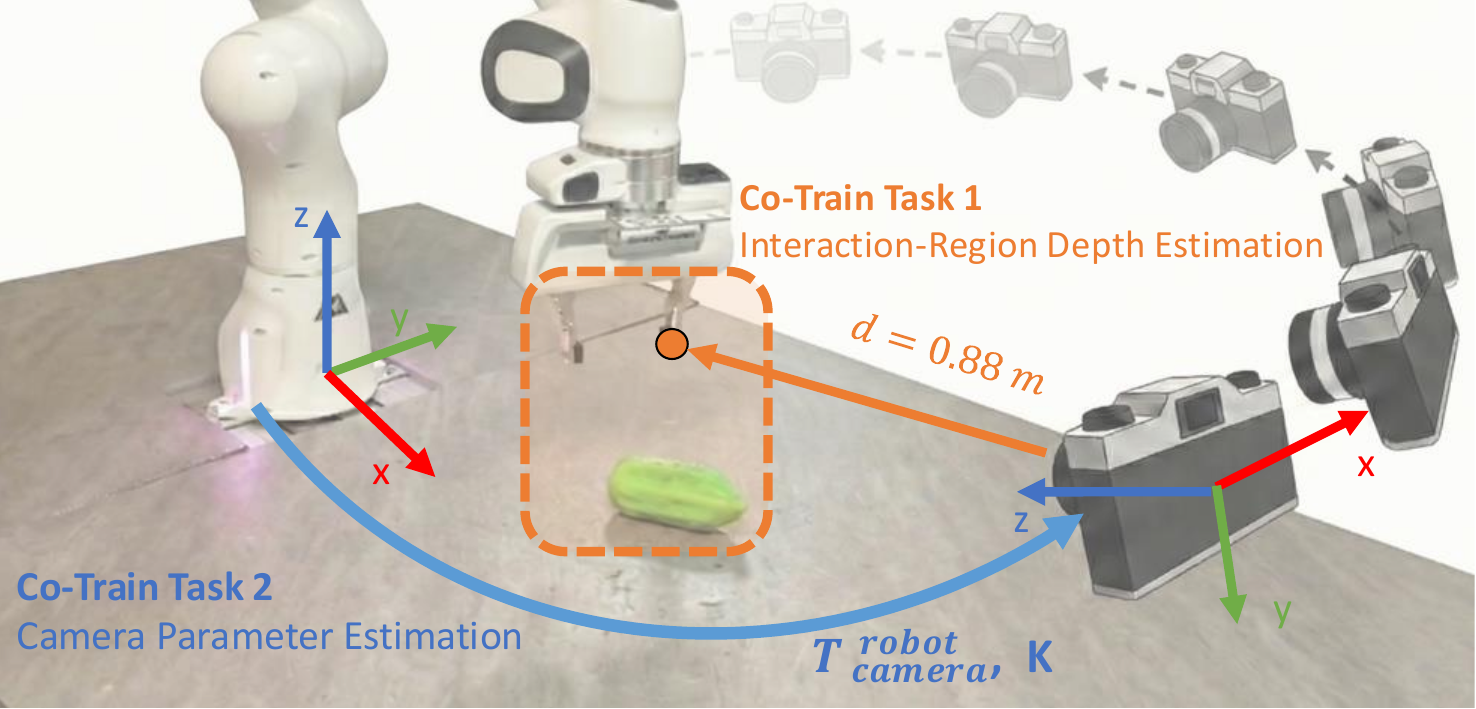}
    \caption{\textbf{Synergistic co-training tasks for viewpoint robustness.} Task 1 performs Interaction-Region Depth Estimation ($d$) for precise spatial reasoning. Task 2 performs Camera Parameter Estimation ($T^{robot}_{camera}, K$) to align perception and action coordinate systems.}
    \label{fig:cotrain}
    \vspace{-0.7cm}
\end{figure}

\subsection{Co-training Tasks for Viewpoint Robustness}
While many existing VLA models operate as ``black boxes'', we decouple the perception-action mapping into camera-frame spatial perception and robot-frame action generation. To improve viewpoint robustness without increasing inference latency, we introduce two synergistic tasks co-trained alongside action chunk prediction, illustrated in Figure~\ref{fig:cotrain}.

\textbf{Interaction-Region Depth Estimation.}
To enhance fine-grained geometric awareness, we utilize an auxiliary task where the model predicts the metric depth $d$ for sampled coordinates $(x, y)$. Rather than uniform sampling, which often captures task-irrelevant background (e.g., walls), we employ task-aware sampling focused on the interaction region. This region, defined by the 2D bounding boxes of the gripper and target object, forces the model to prioritize local spatial relationships, improving precision and training convergence.

Formally, for each sampled $(x,y)$, we query the model with \textit{“What is the depth at ${x,y}$?”} and cast depth prediction as text generation~\cite{qwen3-vl, vla0}. This avoids task-specific regression heads or discretized bins, preserving a unified architecture and leveraging pretrained knowledge. We supervise the generated depth string with cross-entropy loss $\mathcal{L}_{depth}$.

\textbf{Camera Parameter Estimation.}
Our setup features a near-hemispheric orbital camera range, a scope largely underexplored in prior literature, requiring the model to reason about the spatial relationship between the camera and the robot. We supervise the prediction of camera extrinsics $T^{robot}_{camera}$ and the intrinsic $K$ alongside action prediction. 
This co-training strategy encourages camera-aware representation learning under camera pose variations.
The model implicitly strengthens its spatial reasoning capabilities through this co-training task.
Combined with the single-step visual input design, it achieves robustness to viewpoint changes during task execution.

Specifically, the model is queried with the prompt: \textit{“What are the camera extrinsics and intrinsics?”}. It is then required to generate a string representation of the ground-truth parameters, comprising the translation $(x, y, z)$, the Euler rotation $(roll, pitch, yaw)$ in the robot frame, and the intrinsic parameters $(f_x, f_y, c_x, c_y)$. Consistent with our depth estimation approach, we cast the continuous value into plain text and supervise it with a standard cross-entropy loss $\mathcal{L}_{camera}$.

\subsection{Dataset Curation and Sim-to-Real Transfer}
While large-scale real-world robotic datasets, such as Open-X~\cite{oxe} and AgiBot-World~\cite{agibot_world}, often lack stereo image pairs, recent studies (e.g., GraspVLA, InternData-A1, Genie Sim 3, RoboTwin 2.0)~\cite{graspvla, interndata, geniesim3, robotwin} have highlighted the promising scalability of using synthetic data as a primary source for training VLA models. By integrating high-fidelity modern simulation with vision-language foundation models, these works demonstrate robust zero-shot sim-to-real transfer. Inspired by these findings, we curate a large-scale synthetic dataset comprising 5 million stereo trajectories, covering both a table-top Franka and a humanoid robot to evaluate our method's versatility across different robot morphologies.

We mitigate sim-to-real gaps in both appearance and dynamics. For visuals, we adopted three key designs: (i) keep pretrained vision encoders frozen to retain robust foundation features, (ii) co-train on GRIT~\cite{Kosmos2} with 2D box prediction to better align to real imagery, and (iii) use high-fidelity ray-traced rendering with extensive randomization of lighting, textures, and camera parameters. We also observe that while modern rendering achieves high RGB fidelity, it fails to accurately model sensor-specific depth noise. Consequently, stereo vision emerges as a more robust modality for sim-to-real geometric transfer.
For dynamics, we use a quasi-static setting with simple positional control and randomize friction in simulation to improve robustness; dataset details are in the supplementary material.


\subsection{Training Details}
Following GraspVLA, we adopt progressive action generation, where the model first predicts the 2D bounding box of the target object and the next keyframe pose of the gripper as intermediate steps to guide action chunk prediction. Since the left and right stereo images contain largely redundant semantic information, we only require the model to predict the bounding box on the left image to reduce computational overhead. The overall training loss is defined as  
\begin{equation}
\mathcal{L} = \mathcal{L}_{action} + \mathcal{L}_{depth} + \mathcal{L}_{camera} + \mathcal{L}_{bbox} + \mathcal{L}_{pose},
\end{equation}
where $\mathcal{L}_{action}$ is the flow-matching loss for action chunk prediction, and $\mathcal{L}_{depth}$, $\mathcal{L}_{camera}$, $\mathcal{L}_{bbox}$, and $\mathcal{L}_{pose}$ are cross-entropy losses for depth estimation, camera parameter estimation, bounding box prediction, and keyframe pose prediction, respectively.
We balance samples from the five tasks with a ratio of 1:2:2:2:1.
The final model is trained for 300k steps on 32 NVIDIA H800 GPUs with a batch size of 384 and a learning rate of 1.6e-4.

%% file: sec/4_exp.tex
\section{Evaluation}

We conduct comprehensive experiments to answer three key questions:
1) How does \thiswork{} compare to existing VLAs on representative manipulation tasks?
2) How does the stereo camera setting compare to other camera settings under comparable training conditions with state-of-the-art VLAs?
3) What is the individual contribution of our proposed design components to the overall performance?

\subsection{Real-world Experiments\label{sec:main-result}}

\input{fig/main_table}

\subsubsection{Task Suite}

We evaluate our model across six task categories: Common, Distractor, Environment, Transparent Object, Small Object, and Bar-shaped Object tasks. This suite systematically assesses manipulation capabilities, robustness to visual interference, and high-precision execution. We detail each task category below.

\textbf{Common Tasks}. These general tasks involve pick-and-place and stacking operations with daily items (e.g., bowls, cubes). They verify the model's ability to map language instructions to basic motor skills and spatial coordination.

\textbf{Distractor Tasks}. Target objects are placed in cluttered scenes alongside nine irrelevant items of similar appearance. This category evaluates the model's visual grounding and its ability to distinguish target objects from visual distractors.

\textbf{Environment Tasks}. We test generalizability by varying backgrounds and lighting conditions. These tasks evaluate the model's robustness to real-world variations.

\textbf{Transparent Object Tasks}. Focusing on materials like plastic, these tasks provide a detailed comparison with depth-based baselines regarding the handling of objects with special material properties.

\textbf{Small Object Grasping Tasks}. Manipulating small objects ($1\sim2~\mathrm{cm}$) demands millimeter-level precision. These tasks evaluate the fine-grained integration of high-resolution perception and delicate motor control.

\textbf{Bar-shaped Object Grasping Tasks}. Bar-shaped objects like forks have a significantly larger length compared to their width and thickness. Their short axis requires precise estimation of the spatial relationship between the gripper and the object, while the long axis often leads to visual overlap with the gripper in camera views, misleading the model into grasping at improper positions. To systematically evaluate the performance, we test with objects oriented at $0^\circ$, $45^\circ$, and $90^\circ$ relative to the camera.

\subsubsection{Hardware Setup}
\begin{figure}[h]
    \centering
    \includegraphics[width=0.7\linewidth]{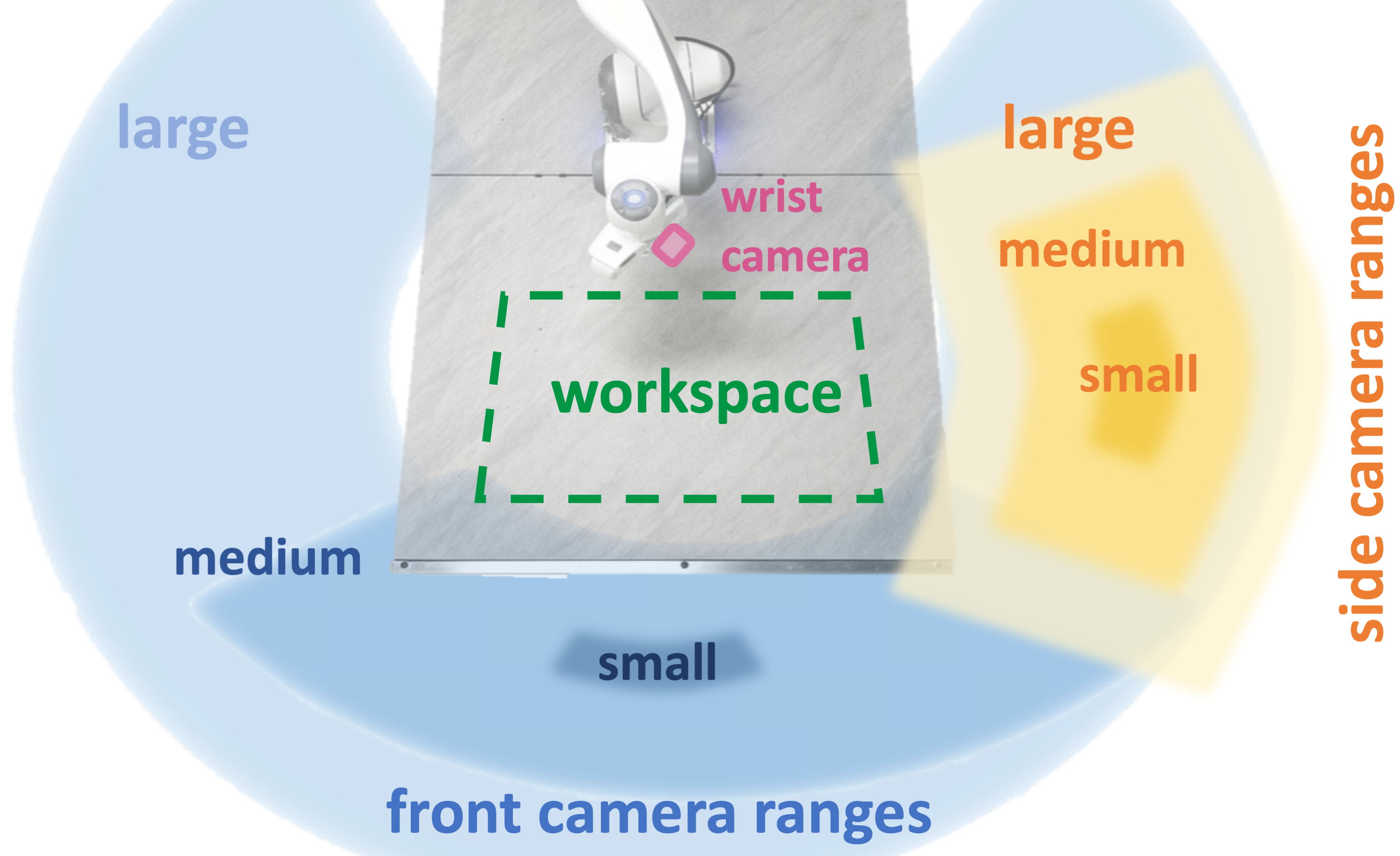}
    \caption{\textbf{Layout of the robot arm, workspace, and cameras.} To evaluate viewpoint robustness, cameras are randomized within three spherical shell regions, ranging from localized sectors to a near-hemispheric volume. While these are 3D spaces, we provide only their top-down projections for visual clarity. See the Appendix for geometric parameters.
    }
    \label{fig:camera-config}
\vspace{-0.7cm}
\end{figure}

The hardware setup is shown in Figure~\ref{fig:camera-config}.
Target objects and distractors are placed in a workspace sizing $0.5m\times0.4m$.
To systematically evaluate the impact of camera settings, we introduce three levels of pose randomization for both front and side cameras, as shown in the figure.
Unless otherwise specified, we employ the small randomization range. Results for all randomization ranges are reported in Section~\ref{sec:eval-camera-settings}.


\subsubsection{Baselines\label{sec:exp-baselines}}
We select baselines to support two complementary comparisons: stereo RGB-based adapted baselines evaluate how existing VLA architectures perform with the same stereo observations, whereas cross-setting baselines compare system-level performance against representative VLAs under their respective camera configurations. For fairness, all baselines are trained or fine-tuned with matched dataset scale and task distribution.

\textbf{Architectural comparison.} Since no existing VLA models natively support stereo inputs, we adapt the aforementioned SOTAs, $\pi_{0.5}$ and GraspVLA, to process synchronized stereo RGB pairs, denoted as \textbf{$\pi_{0.5}$-S} and \textbf{GraspVLA-S}. 
GraspVLA-S is effectively \thiswork{} without our stereo-centric designs, serving as a controlled baseline for testing whether a camera swap alone can explain the gains. To avoid bias toward GraspVLA's original front-side configuration, GraspVLA-S is trained from VLM initialization, while $\pi_{0.5}$-S is fine-tuned using its recommended hyperparameters.

\textbf{Cross-setting comparison.} We compare against representative SOTA models for different modalities. SpatialVLA~\cite{spatialvla} is selected for the RGB-D setup, as it explicitly uses depth as positional embedding to enhance spatial reasoning. For the front-wrist configuration, we employ $\pi_{0.5}$~\cite{pi0.5}, a generalist VLA recognized for its robustness and generalizability. Finally, we adopt GraspVLA~\cite{graspvla} to represent the front-side setup.

\textbf{Depth-based variants}. To systematically assess the impact of depth quality, we address SpatialVLA's~\cite{spatialvla} reliance on potentially inaccurate monocular estimation by introducing two variants: \textbf{SpatialVLA-SD} (Sensor Depth), which utilizes raw depth from the Intel RealSense D435 sensor, and \textbf{SpatialVLA-ED} (Estimated Depth), which employs high-quality depth maps predicted from stereo RGB pairs via FoundationStereo.

\input{tables/camera_setting}

\subsubsection{Metrics}
Task success rates are reported for each test set, with 15 trials per set, yielding a total of \(15\times6\times5=450\) trials. To rigorously assess fine-grained spatial perception and manipulation, we incorporate three additional criteria. First, to prevent models from succeeding through repeated attempts, each trial allows only a single execution: one gripper-close before grasping and one gripper-open after grasping. Second, we disable the ``gripper sticking'' heuristic used in OpenVLA, SpatialVLA, and $\pi_{0.5}$, since it artificially delays gripper actions and can conceal inaccurate decisions. Third, a trial is deemed successful only when the task is fully completed, with no partial credit assigned. While these stricter criteria produce lower success rates than those reported in prior work, they provide a more accurate measure of model capability.

\subsubsection{Results}

Figure~\ref{fig:main-results} shows the success rates across the evaluation tasks.
Our model consistently outperforms all baselines across every task, demonstrating robustness to visual distractors and complex environments. In the rigorous small-size object ($1\sim2~\mathrm{cm}$) grasping task where all baselines completely fail (0\%), our model achieve a 33\% success rate. Performance in this regime is currently limited by the $224 \times 224$ resolution, where small objects span only a few pixels; while higher-resolution backbones could enhance semantic grounding, they introduce a computational trade-off critical for future optimization. 

\textbf{Analysis of depth-based baselines.} Our evaluation of depth-based models shows that performance scales from monocular estimation (lowest) to sensor depth, and finally to stereo estimation (highest). This is likely due to the sim-to-real divergence: simulated depth is noise-free, whereas sensor and monocular data contain significant artifacts. Furthermore, sensors fail on transparent surfaces due to their underlying physical principles, causing the model to struggle with these specific objects. While FoundationStereo~\cite{foundationstereo} produces depth quality on par with simulation, their iterative optimization process results in a 70ms inference overhead. This latency makes them unsuitable for real-time action control.

\subsubsection{Qualitative Analysis}
$\pi_{0.5}$ and the original SpatialVLA model often approached correct target objects, but consistently closed the gripper too early, likely due to the lack of precise spatial perception. 
In contrast, \thiswork{} achieves the highest success rate on all tasks, demonstrating its superior performance. Interestingly, SpatialVLA-ED struggled despite utilizing superior stereo-derived depth. This suggests that the down-sampling of depth maps during spatial encoding may result in the loss of critical geometric details. Meanwhile, GraspVLA-S, which encodes left and right images independently, performed well on bar-shaped objects but failed significantly on more complex tasks. We hypothesize that while GraspVLA's additional pose supervision enhances general spatial awareness, it remains unable to capture fine-grained spatial relationships from the subtle disparities between stereo views.

Overall, these results suggest that the superior performance of \thiswork{} stems from methodological innovations rather than from data scale or quality, emphasizing the critical role of model architectures and training strategies in the stereo setting.

\subsection{Comparison of Camera Settings\label{sec:eval-camera-settings}}

This section provides a system-level comparison of representative VLA methods across common camera settings. Building on the baselines introduced in Sec.~\ref{sec:exp-baselines}, we follow each method's original model design and training protocol, with matched dataset scale and task distribution across methods.
We evaluate four common configurations~\cite{spatialvla,pi0.5,openvla,graspvla,oxe}: single-view RGBD (SpatialVLA-ED), stereo RGBs (\thiswork{}), front + wrist RGBs (fine-tuned $\pi_{0.5}$), and front + side RGBs (GraspVLA).

To evaluate the robustness to camera pose variations, we generate equal-sized datasets across three randomization ranges (Figure~\ref{fig:camera-config}) and train/fine-tune each model accordingly. To further account for potential motion blur induced by camera movement and vibrations, we include a 'moving' test track where the camera traverses from left to right relative to the robot. In all evaluations, we maintain the gripper and objects within the frame to ensure they are visible. The randomization ranges are defined as spherical shells centered on the workspace. For the front view:
\begin{itemize}
    \item The \textbf{small range} corresponds to an elevation range of $20^\circ \sim 30^\circ$, an azimuth range of $-10^\circ \sim 10^\circ$, and a radius range of $1.2\,\text{m} \sim 1.3\,\text{m}$.
    \item The \textbf{large range} covers an elevation of $15^\circ \sim 45^\circ$ and a radius of $0.9\,\text{m} \sim 1.6\,\text{m}$. Instead of a full $2\pi$ azimuth, we restrict the range to $-150^\circ \sim 150^\circ$ to exclude the region directly behind the robot, thereby avoiding self-occlusion caused by the robotic arm.
\end{itemize}
Detailed geometric parameters are provided in the Appendix, and average task success rates are reported in Table~\ref{table:compare-camera-views}. While the front + side configuration with GraspVLA performs well under small randomization,
it suffers a significant drop in larger randomization due to the difficulty of extracting coherent spatial cues from unaligned views. In this evaluation, \thiswork{} obtains the highest success rates across the tested randomization ranges, indicating robustness to camera pose variations under the stereo setup.

Regarding the ``front + wrist'' configuration ($\pi_{0.5}$), although it starts with a lower baseline under small randomization, it exhibits higher resilience to increasing pose variations. This is because the wrist camera provides egocentric views of gripper-object interactions that remain invariant to global camera shifts. However, this setup is prone to premature gripper closure, a failure mode likely caused by the wrist camera's limited precise spatial perception. Finally, RGBD-based models consistently underperform, particularly in large randomization, likely due to the lack of current patch-level depth encoding methods in capturing fine-grained spatial cues.

We believe that each camera setting has potential depending on the deployment scenario. Within this context, our results suggest that the stereo configuration is a promising alternative for balancing task performance, robustness to camera pose variation, and deployment simplicity.

\input{tables/depth_humanoid}

\subsection{Ablation of Key Design Components\label{sec:ablation-components}}

\input{tables/component_ablation}

Table~\ref{table:component-ablation} summarizes the component-wise ablation. We use GraspVLA-S, introduced in Sec.~\ref{sec:exp-baselines}, as the stereo-only baseline and progressively add GeoSem, DE, and CPE to isolate their contributions. The stereo-only baseline achieves moderate performance, indicating that a camera swap alone cannot explain the overall gains. GeoSem and DE provide substantial improvements by enhancing stereo geometry encoding and interaction-region spatial supervision, respectively. CPE further improves robustness under large and moving camera pose shifts, leading to the strongest overall performance.

\subsection{Design Choices of \featurefusion{}\label{sec:ablation-feature}}

We conduct ablation experiments on the \featurefusion{} module to evaluate the effectiveness of our feature selection and fusion strategies.

For the selection of FoundationStereo features, we evaluate models in simulation after 100k training steps, averaging the performance of the three most recent checkpoints at 10k-step intervals.
As shown in Table~\ref{table:ablate-featurefusion}, success rates progressively improve when using $V_{corr}$, $V_{c}$, and $V_c'$, respectively. This supports our hypothesis that $V_{corr}$ retains little geometric information, whereas the unfiltered cost volume $V_{c}$ lacks the long-range spatial relationships present in the filtered version $V_c'$. The superior performance of $V_c'$ validates our choice of this representation as the stereo feature.

In addition, the results show that incorporating the semantic feature consistently yields higher success rates than using the stereo feature alone. Manual inspection of the rollouts revealed that models without the semantic feature more frequently grasped incorrect objects, indicating that semantic features is essential for vision-language grounding.
This justifies the necessity of fusing both features.

As is shown in Figure~\ref{fig:feature_fusion_results}, for feature fusion strategies, concatenating features by sequence yields slightly lower success rates and increases computational overhead due to the doubled number of visual tokens. Channel-wise concatenation provides both efficiency and performance advantages.

In addition to FoundationStereo~\cite{foundationstereo}, other foundation models, such as VGGT~\cite{vggt}, can be alternatives for stereo feature extraction.
We compare the depth estimation precision of VGGT and FoundationStereo on our dataset to assess their performance in manipulation scenarios. VGGT obtains an AbsRel~\cite{absrel} error of 0.4121, whereas FoundationStereo achieves a substantially lower error of 0.0275. These results indicate that FoundationStereo provides more reliable depth information, likely due to its specialization in stereo inputs. In contrast, VGGT may excel in broader multi-view settings without stereo assumptions.

\subsection{Effectiveness of Co-training Tasks\label{sec:ablation-cotrain}}

\subsubsection{Effectiveness of Interaction-Region Depth Estimation}
We compare two alternatives: predicting the depth of points uniformly sampled in the whole image, and no prediction of depth.
As shown in Figure~\ref{fig:depth_est_results},
\depthestimation{} facilitates model learning and achieves the highest performance. In contrast, predicting depth over the entire image negatively impacts training in the early stages, potentially due to the additional burden of learning irrelevant background depths that do not contribute to manipulation success. Nevertheless, both strategies stabilize training in later stages. This suggests that explicitly supervising the model to perceive critical spatial details is beneficial.

\subsubsection{Effectiveness of Camera Parameter Estimation}
Quantitative results in Table~\ref{table:component-ablation} indicate that this task substantially improves viewpoint robustness, with the most significant gains observed in the large and moving categories. The model's stability under moving camera conditions is notable. We hypothesize this stems from the synergy between viewpoint randomization during training and the model’s dependence solely on current-step observations.

\subsection{Fine-Tuning on a Humanoid Platform}
Motivated by the similarity between stereo vision and human visual perception, we extend our analysis to a humanoid platform and evaluate performance on two household tasks: Cleaning and Laundry. Using 500 trajectories for fine-tuning and 15 trials for evaluation, the results in Table~\ref{table:crossemb_finetuning} demonstrate that pre-training significantly improves downstream adaptation. However, Laundry achieves a lower success rate than Cleaning, which we attribute to the distributional shift from the table-top-only pre-training data. See the Appendix for details.

\subsection{Inference Speed}

We profiled the model to quantify inference time overhead of each component on a NVIDIA H800 GPU. A full FoundationStereo forward pass to predict depth takes 68 ms per iteration, while extracting the filtered cost volume $V_c'$ requires only 14 ms, reducing computation by 54 ms. We therefore use $V_c'$ as the stereo feature representation, yielding a vision-encoder inference time of 27 ms.

\input{tables/inference_speed}

%% file: fig/main_table.tex
\begin{figure*}[t]
    \centering
    \includegraphics[width=\linewidth]{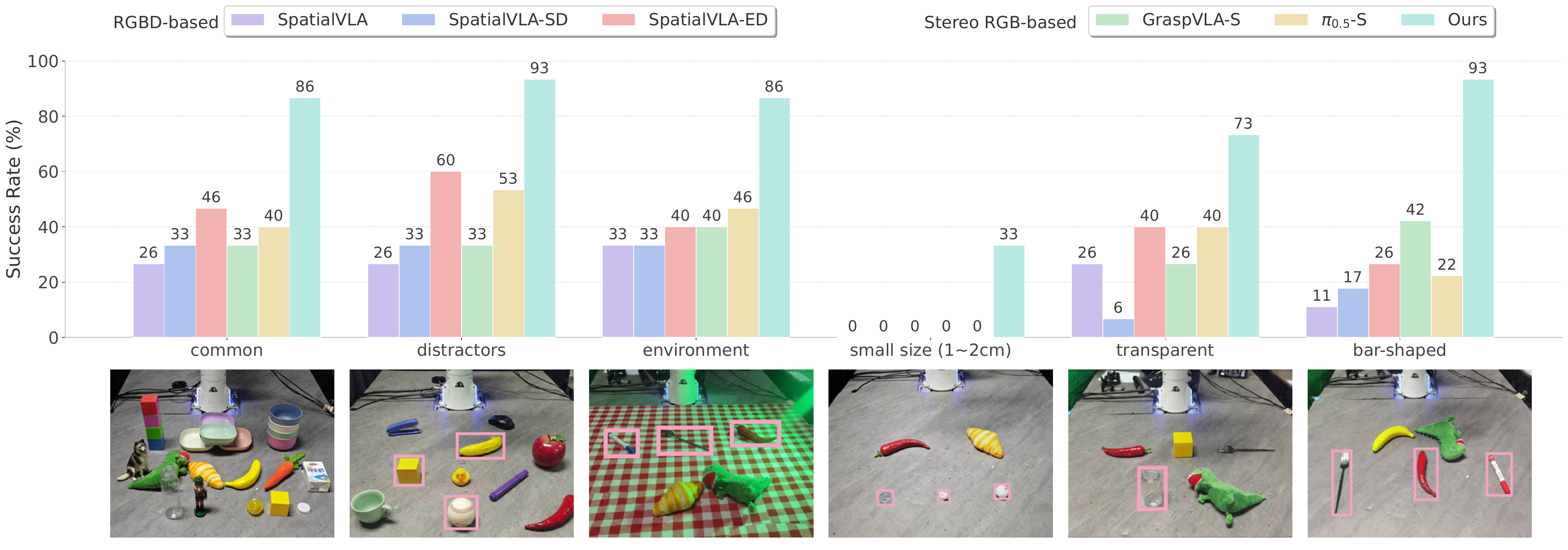}
    \caption{\textbf{Real-world evaluation tasks and results.} 
We evaluate \thiswork{} on a comprehensive suite of tasks requiring fine-grained perception, manipulation and generalizability. Objects enclosed by colored boxes denote the target objects used for evaluation, while other items in the scene serve as distractors. 
We include RGB-D-based baselines to compare system-level performance with representative depth-aware VLA systems, and Stereo RGB-based baselines to evaluate how different architectures exploit the same stereo observations. All models are trained on an identical synthetic dataset. \thiswork{} consistently outperforms baselines across the evaluated tasks, demonstrating its effectiveness.}
    \label{fig:main-results}
\end{figure*}

%% file: tables/camera_setting.tex
\begin{figure*}[t]
\begin{minipage}[t]{0.7\textwidth}
\resizebox{1.0\textwidth}{!}{
\begin{tabular}{lllrrrr} 
\toprule
modalities & viewpoint & model  & small  & medium & large & moving \\
\midrule
RGBD & front & SpatialVLA-ED~\cite{spatialvla}  & 35.6{\small$\pm$6.9}\% & 6.7{\small$\pm$3.7}\% & 3.9{\small$\pm$3.0}\% & 2.8{\small$\pm$2.6}\% \\
RGBs & front + wrist & $\pi_{0.5}$~\cite{pi0.5}  & 65.6{\small$\pm$6.9}\% & \underline{51.7{\small$\pm$7.2}\%} & \underline{42.2{\small$\pm$7.1}\%} & \underline{39.4{\small$\pm$7.1}\%} \\
RGBs & front + side & GraspVLA~\cite{graspvla} & \underline{71.7{\small$\pm$6.5}\%} & 23.3{\small$\pm$6.1}\%  & 11.1{\small$\pm$4.6}\%  & 7.2{\small$\pm$3.8}\% \\
stereo RGBs & front & StereoVLA  & \textbf{77.2{\small$\pm$6.1}\%} & \textbf{62.2{\small$\pm$7.0}\%}  & \textbf{53.9{\small$\pm$7.2}\%} & \textbf{52.2{\small$\pm$7.2}\%}\\
\bottomrule
\end{tabular}
}
\captionof{table}{\textbf{Comparison of camera settings.} 
StereoVLA demonstrates strong robustness to camera pose variations. To ensure fairness, all models are trained on the same amount of synthetic data for each camera randomization range. Section \ref{sec:eval-camera-settings} details the experimental setup. Best results are \textbf{bolded}, second-best results are \underline{underlined}.
Each reported success rate is estimated from three evaluation groups, each covering 6 tasks with 10 trials per task, for 180 real-world robot trials in total. The $\pm$ values denote 95\% Wilson confidence intervals.
}
\label{table:compare-camera-views}
\end{minipage}
\hfill
\begin{minipage}[t]{0.27\textwidth}
\centering
\resizebox{0.9\textwidth}{!}{
\begin{tabular}{rrr}
\toprule
feature                  & w/o sem. & w/ sem. \\ 
\midrule
$V_{corr}$ & 27.0\%    & 54.0\%                       \\
$V_c$  & 40.0\%    & 69.0\%                       \\
$V_c'$   & 51.0\%    & \textbf{77.0\%}              \\
\bottomrule
\end{tabular}
}
\captionof{table}{\textbf{Comparison of stereo feature selections.} The adoption of the filtered cost volume $V_c'$ and the semantic feature results in the highest success rate.}
\label{table:ablate-featurefusion}
\end{minipage}
\vspace{-0.7cm}
\end{figure*}

%% file: tables/depth_humanoid.tex
\begin{figure*}[t]
\begin{minipage}[t]{0.32\textwidth}
\centering
\vspace{-3.8cm}
\includegraphics[width=\linewidth]{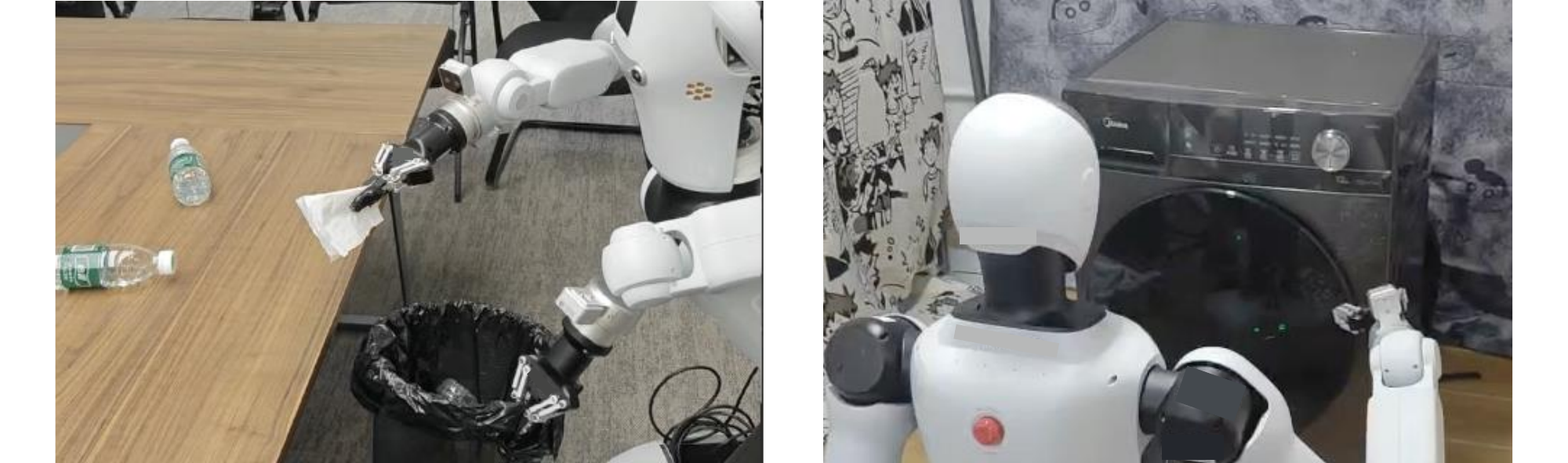}
\par
\vspace{0.4cm}
\begin{tabular}{rrr}
\toprule
Model   & Cleaning  & Laundry \\ 
\midrule
w/o pre-train & 26.6\%    & 13.3\%      \\
w/ pre-train  & \textbf{73.3\%}    & \textbf{53.3\%}      \\
\bottomrule
\end{tabular}
\captionof{table}{\textbf{Humanoid Fine-Tuning.} The model with pre-training demonstrates notably higher success rates in both scenarios.}
\label{table:crossemb_finetuning}

\end{minipage}
\hfill
\begin{minipage}[t]{0.32\textwidth}
\centering
\includegraphics[width=\linewidth]{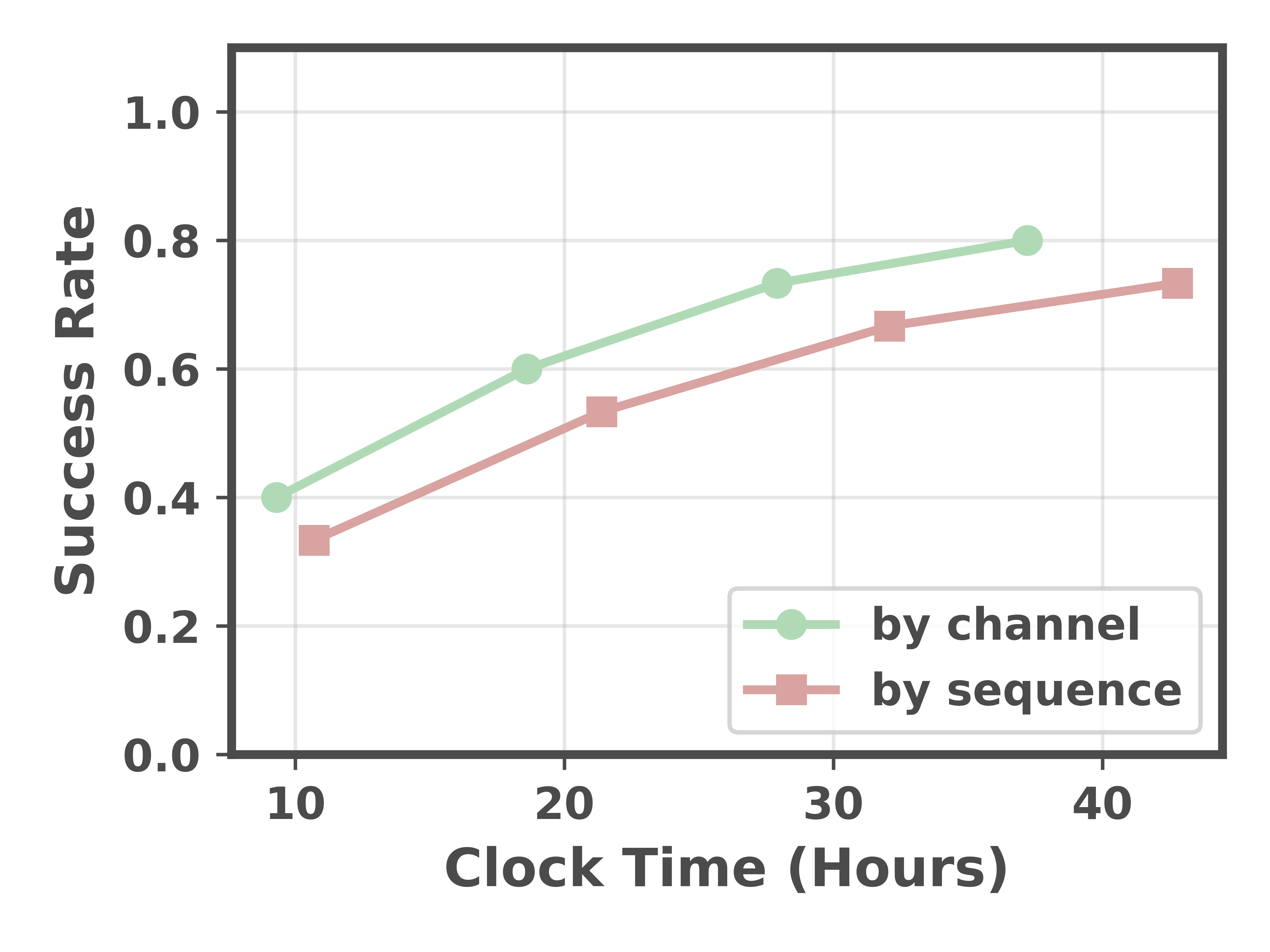}
\caption{\textbf{Comparison of feature fusion methods.} Sequence concatenation results in a lower success rate and requires a longer training time.}
\label{fig:feature_fusion_results}
\end{minipage}
\hfill
\begin{minipage}[t]{0.32\textwidth}
\centering
\includegraphics[width=\linewidth]{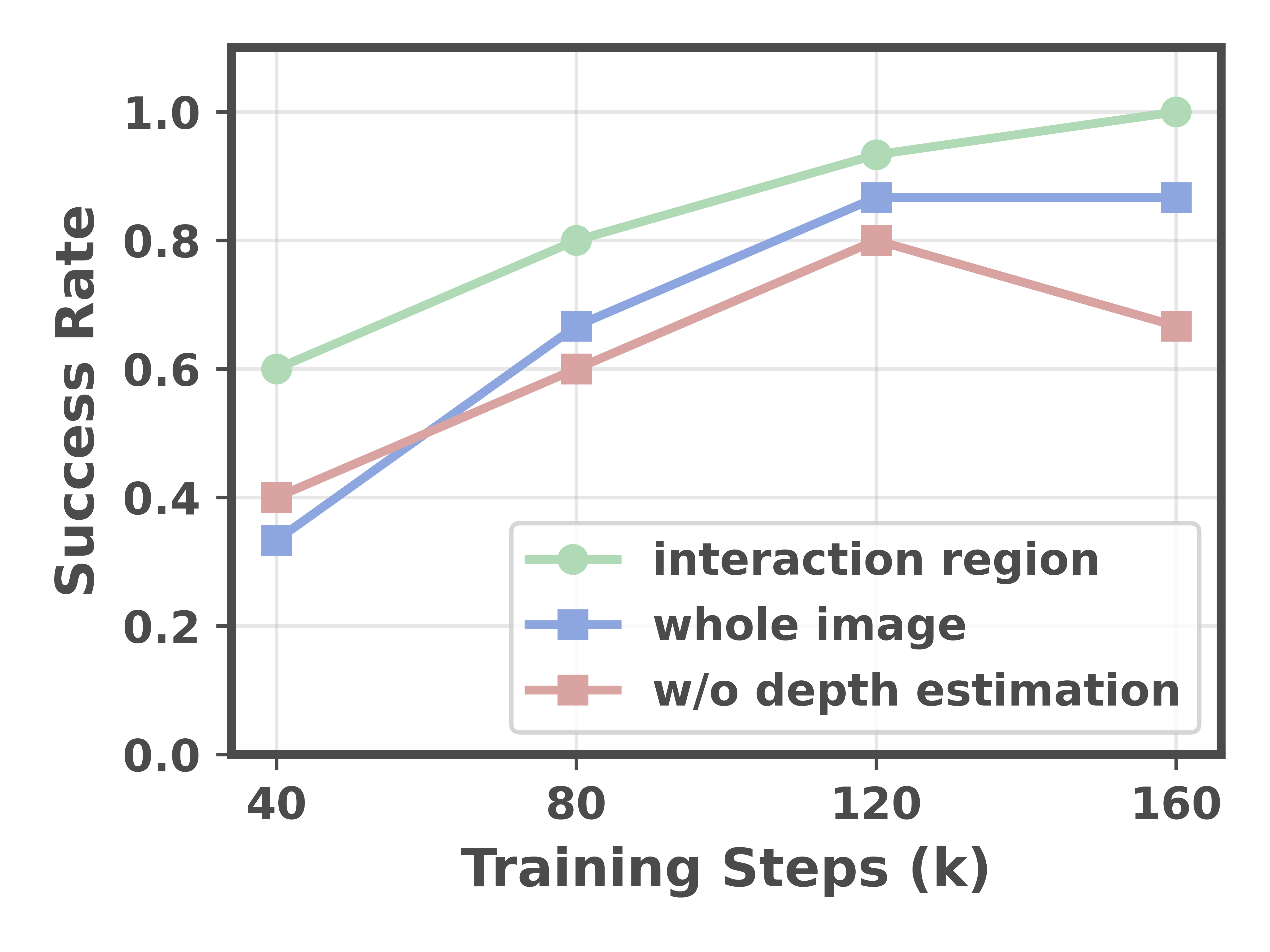}
\caption{\textbf{Comparison of depth estimation methods.} Depth estimation within the interaction region yields the highest success rate.}
\label{fig:depth_est_results}
\end{minipage}
\vspace{-0.5cm}
\end{figure*}

%% file: tables/component_ablation.tex
\begin{table}[htbp]
\centering
\resizebox{0.48\textwidth}{!}{
\begin{tabular}{cccc|cccc}
\toprule
Stereo & GeoSem & DE & CPE & small & medium & large & moving \\ \midrule
\checkmark & & & & 29.0 & - & - & - \\
\checkmark & \checkmark & & & 57.8 & 36.7 & 13.3 & 10.0 \\
\checkmark & \checkmark & \checkmark & & 74.9 & 51.8 & 30.2 & 27.1 \\
\checkmark & \checkmark & \checkmark & \checkmark & \textbf{77.3} & \textbf{62.1} & \textbf{54.1} & \textbf{52.3} \\
\bottomrule
\end{tabular}
}
\caption{\textbf{Ablation of key design components.} DE refers to interaction-region Depth Estimation. CPE refers to Camera Parameter Estimation.}
\label{table:component-ablation}
\vspace{-0.5cm}
\end{table}

%% file: tables/inference_speed.tex
\begin{table}[htbp]
    \centering
    \resizebox{0.42\textwidth}{!}{
    \begin{tabular}{rrrrr}
    \toprule
    & vision & language & action & total \\ 
    \midrule
    time (ms) & 27 & 81 & 52  & 168 \\
    \bottomrule
    \end{tabular}
    }
    \captionof{table}{\textbf{Breakdown on inference time.}}
    \label{table:inference-speed}
    \vspace{-0.5cm}
\end{table}

%% file: sec/5_limitation.tex
\section{Limitation and Future Work}
Despite promising results, our work has two main limitations. First, the input resolution of 224×224 constrains performance on small objects; exploring higher resolutions while balancing computational cost is an important direction. Second, the model does not yet capture long-horizon temporal dependencies, which limits its ability to solve more complex, multi-stage tasks.


%% file: sec/6_conclusion.tex
\section{Conclusion}

We introduced \thiswork{}, a vision-language-action model that exploits rich geometric cues from stereo vision to achieve robust and precise manipulation. Experiments demonstrate that our approach outperforms prior methods by a large margin in the stereo setting and is robust to camera pose variations. Furthermore, our comparison of camera setups provides practical guidance for balancing task performance and deployment overhead in future studies.

%% file: sec/7_supp.tex
\section{Zero-shot Evaluation on \benchmark}

\input{tables/libero}

\noindent{\textbf{\benchmark: Robustness Evaluation under Extended Randomization.}} To conduct a more rigorous assessment of model generalization, we extend the widely-adopted LIBERO benchmark~\cite{libero} by introducing additional environmental variations. Drawing inspiration from the randomization protocols in 4D-VLA~\cite{4dvla} and LIBERO-Pro~\cite{liberopro}, we implement the following two enhancements specifically designed to challenge the models' adaptability beyond standard scenarios:

\begin{enumerate}
    \item \textbf{Camera Viewpoint Shifts:} This modification is made to help evaluate the model's robustness to camera viewpoint generalization. Inspired by MVBench~\cite{4dvla}, we test each policy across four orientations: $0^\circ, 45^\circ, 90^\circ,$ and $135^\circ$. For OpenVLA and $\pi_{0.5}$, we perform an azimuthal rotation on the agent-view camera around a fixed look-at point. For StereoVLA, we perform an azimuthal rotation on the left camera of the stereo pairs around a fixed look-at point, and then derive the extrinsics of the right camera correspondingly, thereby maintaining a consistent stereo baseline and geometric relationship.
    \item \textbf{Object Position Perturbations:} By decoupling specific objects from fixed spatial regions, we effectively prevent the model from achieving success via rote memorization. Drawing inspiration from LIBERO-Pro~\cite{liberopro}, we randomly shuffle the assignments between objects and their predefined initialization regions. Furthermore, we apply a stochastic perturbation of $\pm 10~cm$ along both the $x$ and the $y$ axes to the initial position of the basket.
\end{enumerate}

To get a taste of \benchmark, see Figure~\ref{fig:gallery}, where we present a gallery of representative task executions in the test suite.

\input{fig/gallery_image.tex}

\begin{figure}[h]
    \centering
    \includegraphics[width=0.9\linewidth]{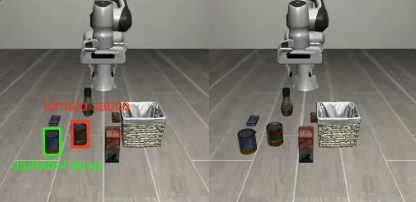}
    \caption{\textbf{Visual similarity between certain objects in LIBERO test suite.} Objects such as ``alphabet soup'' (highlighted in \textcolor{green}{green}) and ``tomato sauce'' (highlighted in \textcolor{red}{red}) exhibit extreme visual similarity, making them nearly indistinguishable. This inherent ambiguity motivates our use of relaxed success criteria and category-aware instructions to ensure a fair assessment for zero-shot models.
    }
    \label{fig:visual-similarity}
\end{figure}

\textbf{Refined success criteria and standardized language instructions.} The standard LIBERO-Object test suite poses a significant challenge for zero-shot models due to inherent ambiguities in target object descriptions. As illustrated in Figure~\ref{fig:visual-similarity}, the visual similarity between certain items (such as ``alphabet soup'' and ``tomato sauce'') makes them almost indistinguishable, even to human observers. To address these problems, we relax the success criteria to reflect category-level completion, and modify the language instructions by replacing fine-grained, category-ambiguous labels with category-aware descriptions.


\textbf{Setup.} We evaluate the models on \benchmark. A trial is considered successful if the robot successfully grasps and moves the target object (or any object belonging to the same semantic category) to the containing region of the basket. For each model under a specific camera viewpoint ($0^\circ, 45^\circ, 90^\circ, 135^\circ$), success rates are computed based on performances across 10 distinct tasks, with 50 trials per task, yielding a total of \(50\times10=500\) trials.


\begin{table}[ht]
    \centering
    \renewcommand{\arraystretch}{1.0}
    \setlength{\tabcolsep}{6pt}
    \begin{tabular}{ccc}
    \toprule
    \textbf{Method} & \textbf{Camera Input} & \textbf{Fine-tuned} \\ \midrule
    OpenVLA-LIBERO & \makecell[c]{third-person view RGB*} & \cmark \\ \midrule
    $\pi_{0.5}$-LIBERO & \makecell[c]{third-person view RGB*\\ + wrist RGB} & \cmark \\ \midrule
    StereoVLA (Ours) & \makecell[c]{third-person view stereo RGBs*} & \xmark \\ \bottomrule
    \addlinespace[1ex]
    \multicolumn{3}{c}{* indicates that the camera is subject to viewpoint shifts}
    \end{tabular}
    \captionof{table}{\textbf{Details of all the methods.} To ensure fair comparison, we compare our model with the official OpenVLA and $\pi_{0.5}$ models fine-tuned on LIBERO. Following their recommendations, OpenVLA uses a third-person view, while $\pi_{0.5}$ integrates both third-person and wrist views. Our model utilizes third-person stereo RGB.}
    \label{table:details-of-methods}
\end{table}

\textbf{Baselines.} For baseline comparisons, we use the officially released checkpoints for OpenVLA-LIBERO~\cite{openvla} and $\pi_{0.5}$-LIBERO~\cite{pi0.5}. These models are fine-tuned on the LIBERO demonstration dataset, curated by OpenVLA to filter out static frames and failure trajectories, and rendered in high resolution. Despite that, as we apply Camera Viewpoint Shifts and Object Position Perturbations in \benchmark, the evaluation on the baseline models can also be regarded as zero-shot. In accordance with the original configurations, OpenVLA-LIBERO utilizes a third-person view RGB camera, whereas $\pi_{0.5}$-LIBERO integrates a multi-view input from both third-person and wrist view RGB cameras. Details of all the methods are presented in Table~\ref{table:details-of-methods}.

\textbf{Analysis of results.} As shown in Table~\ref{table:libero-object-sv}, our model showcases outstanding zero-shot performance on LIBERO-Object-SV, outperforming the baseline models by a significant margin. The performance of baseline models proves to be very fragile under varying camera viewpoints. We observe a performance collapse as the viewpoint angle increases from $0^\circ$ to $135^\circ$, largely because these models are overfitted to the specific visual perspectives of the demonstration data. Moreover, a closer inspection of the failure cases of the baseline models reveals a clear tendency toward spatial overfitting. Specifically, there are some cases where the robot attempts to grasp at the object's initial location before perturbation, instead of tracking its perturbed position in the current scene. This suggests that these models are likely to have memorized the absolute distributions of the objects in the training set, rather than learning genuine spatial reasoning. In contrast, StereoVLA maintains robust performance across camera viewpoint shifts and object position perturbations. We attribute this to the combination of our GeoSem feature extractor and our two synergistic co-training tasks, which enable the model to accurately localize targets in 3D space regardless of the camera's perspective.

\section{Details of Camera Viewpoint Randomization}

The detailed configurations of camera viewpoint randomization settings employed for data generation are summarized in Table~\ref{table:camera-randomization}. For all experiments with stereo input, we use the widely-adopted Zed Mini camera.

\begin{table}[ht]
    \centering
    \renewcommand{\arraystretch}{1.2}
    \setlength{\tabcolsep}{8pt}
    \begin{tabular}{lcc}
    \hline
    \hline
    Randomization Setting & small & large \\ \hline
    Camera to robot angle & $\pm 10^\circ$ & $\pm150^\circ$ \\
    Camera elevation angle & $20^\circ\sim30^\circ$ & $15^\circ\sim45^\circ$ \\
    Camera roll angle & $\pm 5^\circ$ & $\pm 15^\circ$ \\
    Camera look-at offset ($x$) & $\pm 5~cm$ & $\pm 20~cm$ \\
    Camera look-at offset ($y$) & $\pm 5~cm$ & $\pm20~cm$ \\
    Camera look-at offset ($z$) & $0~cm$ & $\pm10~cm$ \\
    Distance to look-at point & $1.2\sim1.3~m$ & $0.9\sim1.6~m$ \\
    Stereo baseline & $6.3~cm$ & $6.3\pm 0.5~cm$ \\
    \hline
    \hline
    \end{tabular}
    \captionof{table}{\textbf{Parameters of camera viewpoint randomization.}}
    \label{table:camera-randomization}
\end{table}

\begin{figure*}
    \centering
    \includegraphics[width=0.8\linewidth]{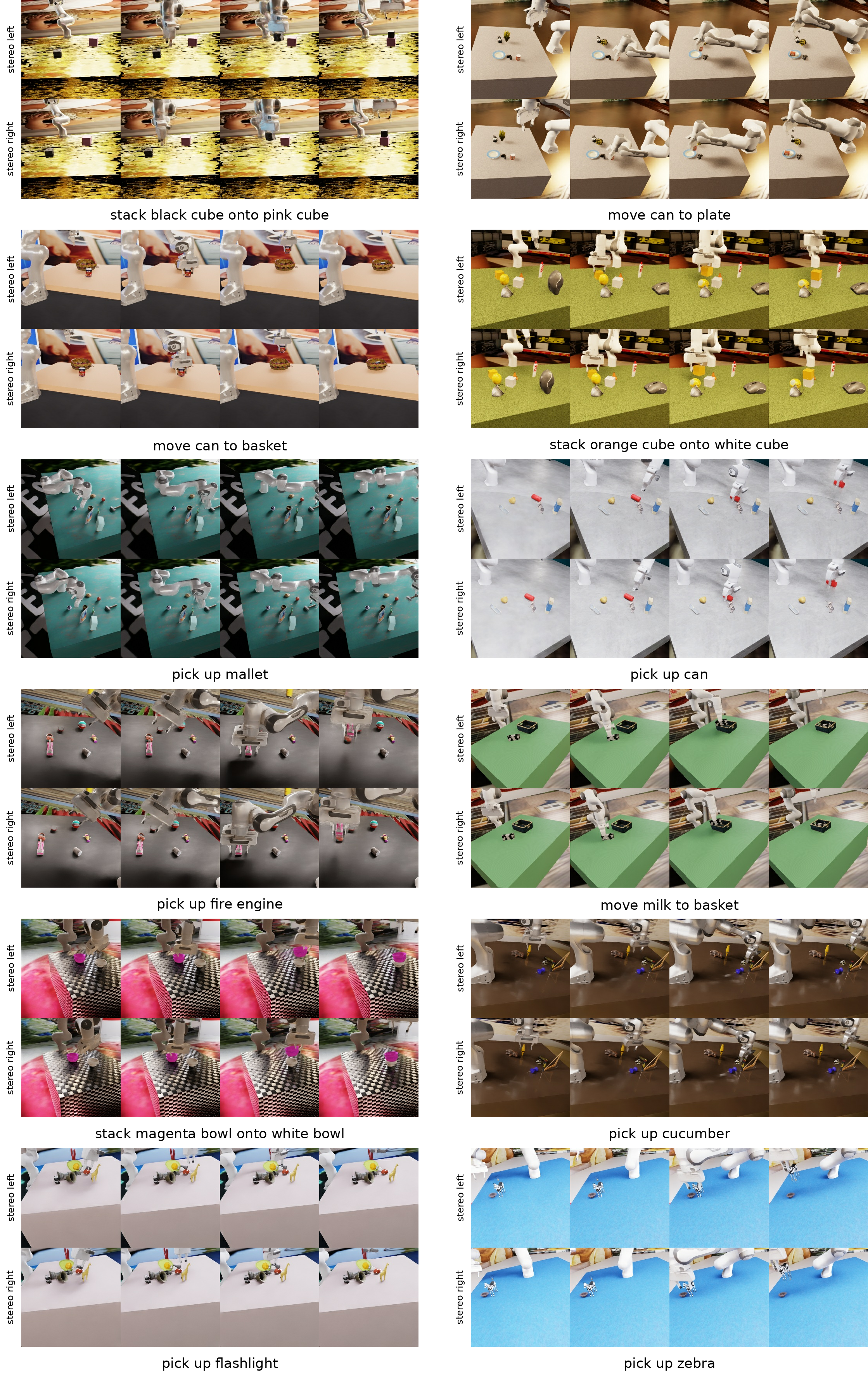}
    \caption{\textbf{Gallery of our training dataset.} 12 randomly sampled trajectories from our synthetic stereo data. For clarity, 4 frames are uniformly sampled from each trajectory for display.}
    \label{fig:dataset}
\end{figure*}

\section{Dataset Details}
\subsection{Mitigation of Sim-to-Real Gap}
Inspired by recent advancements in large-scale robotic learning~\cite{graspvla, robotwin, interndata, geniesim3}, we synthesized a dataset comprising 5 million stereo image pairs. Following established paradigms, we decouple the sim-to-real gap into distinct visual and physical components and implement targeted strategies to mitigate each.

\textbf{Visual Domain Alignment.} The discrepancy between synthetic and real-world RGB inputs has been significantly narrowed by the advent of high-fidelity ray-tracing and pre-trained vision foundation models. Leveraging extensive texture libraries, we generated a highly diverse training set. Consistent with prior findings~\cite{simandrealcotraining}, we observed that even "unrealistic" synthetic materials contribute to a broader training distribution. This diversity ensures the model encapsulates the variance of real-world conditions, thereby enhancing its robustness.

\textbf{Physical Dynamics Modeling.} Discrepancies in physical behaviour typically arise from inaccuracies in contact modeling and material properties. To circumvent the complexities of intricate dynamics, we adopted a quasi-static assumption, utilizing simplified end-effector positional control and binary gripper states. Furthermore, we employed Domain Randomization (DR) across both object and robot material properties to improve generalization.

While these methodologies effectively bridge the gap for prehensile tasks, their efficacy may diminish in contact-rich scenarios, such as non-prehensile manipulation. Investigating these complex physical interactions remains a pivotal direction for future research.

\subsection{Dataset Composition}
To endow the model with generalizable manipulation capabilities, we curated a synthetic dataset encompassing six distinct task categories: object grasping, container-based placement, color-conditioned picking and placing, and the sequential stacking of bowls and cubes. Although the current scope is restricted to prehensile manipulation, these tasks serve as a rigorous testbed for evaluating the model's spatial and geometric reasoning; future work can extend this framework to include more intricate non-prehensile scenarios. While the Franka Panda remains the primary experimental platform, we also integrated a specialized subset of stereo-vision data (500k trajectories) from humanoid robots to assess the method's adaptability across heterogeneous hardware. Our preliminary findings indicate that pre-training on these fundamental tabletop pick-and-place skills significantly facilitates the model's ability to generalize to more complex tasks.

\section{Fine-tuning on a Humanoid Platform}
\begin{figure}[htbp]
    \centering
    \includegraphics[width=0.5\linewidth]{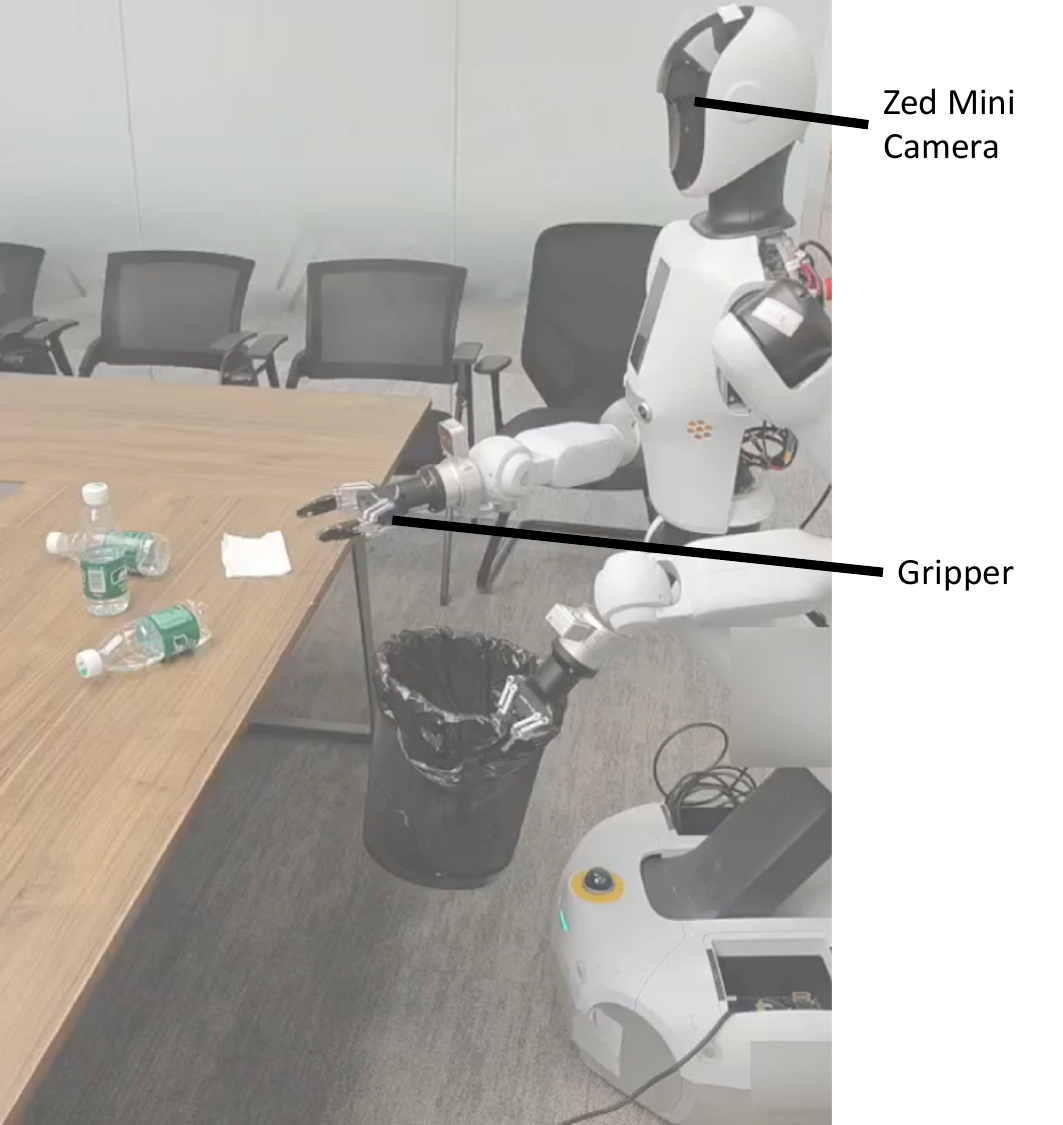}
    \caption{\textbf{Hardware setup of our humanoid experiment.}
    }
    \label{fig:humanoid_setup}
\end{figure}
The experimental setup, shown in Figure~\ref{fig:humanoid_setup}, utilizes the humanoid's head-mounted stereo camera for input. To align with our single-arm pre-training configuration, the model actuates only the right arm while the left arm remains fixed. We evaluate performance on two tasks: Laundry (opening a washing machine) and Cleaning (removing rubbish, such as tissue or empty bottles, from a conference table and dropping it into a bin). As this study focuses on geometric reasoning, we evaluate exclusively on single-step manipulations. We conduct 15 trials per task, measuring performance based on binary success. Both models are fine-tuned for 5000 steps using a batch size of 24 and a learning rate of $1.6 \times 10^{-4}$.

\section{Details of Main Experiments}
For all the methods, we use the recommended hyperparameters in their official repositories to train/fine-tune their models and provide them in Table \ref{table:hyperparams}.

\input{tables/all_methods}




%% file: tables/libero.tex


\begin{table}[ht]
    \centering
    \renewcommand{\arraystretch}{1.3}
    \setlength{\tabcolsep}{10pt}
    \begin{tabular}{lcccc}
    \toprule
    Viewpoint Angle & $0^\circ$ & $45^\circ$ & $90^\circ$ & $135^\circ$ \\ \midrule
    OpenVLA-LIBERO & 9.8\% & 0.2\% & 0.0\% & 0.0\% \\
    $\pi_{0.5}$-LIBERO & 35.4\% & 36.0\% & 15.8\% & 6.8\% \\
    StereoVLA (Ours) & \textbf{65.8\%} & \textbf{59.2\%} & \textbf{51.8\%} & \textbf{38.8\%} \\
    \bottomrule
    \end{tabular}
    \captionof{table}{\textbf{Zero-shot evaluation on \benchmark.} Our model demonstrates superior zero-shot performance on \benchmark, an enhanced version of LIBERO-Object~\cite{libero} with Camera Viewpoint Shifts and Object Position Perturbations.}
    \label{table:libero-object-sv}
\end{table}



%% file: fig/gallery_image.tex
\begin{strip}
    \centering
    \includegraphics[width=\textwidth]{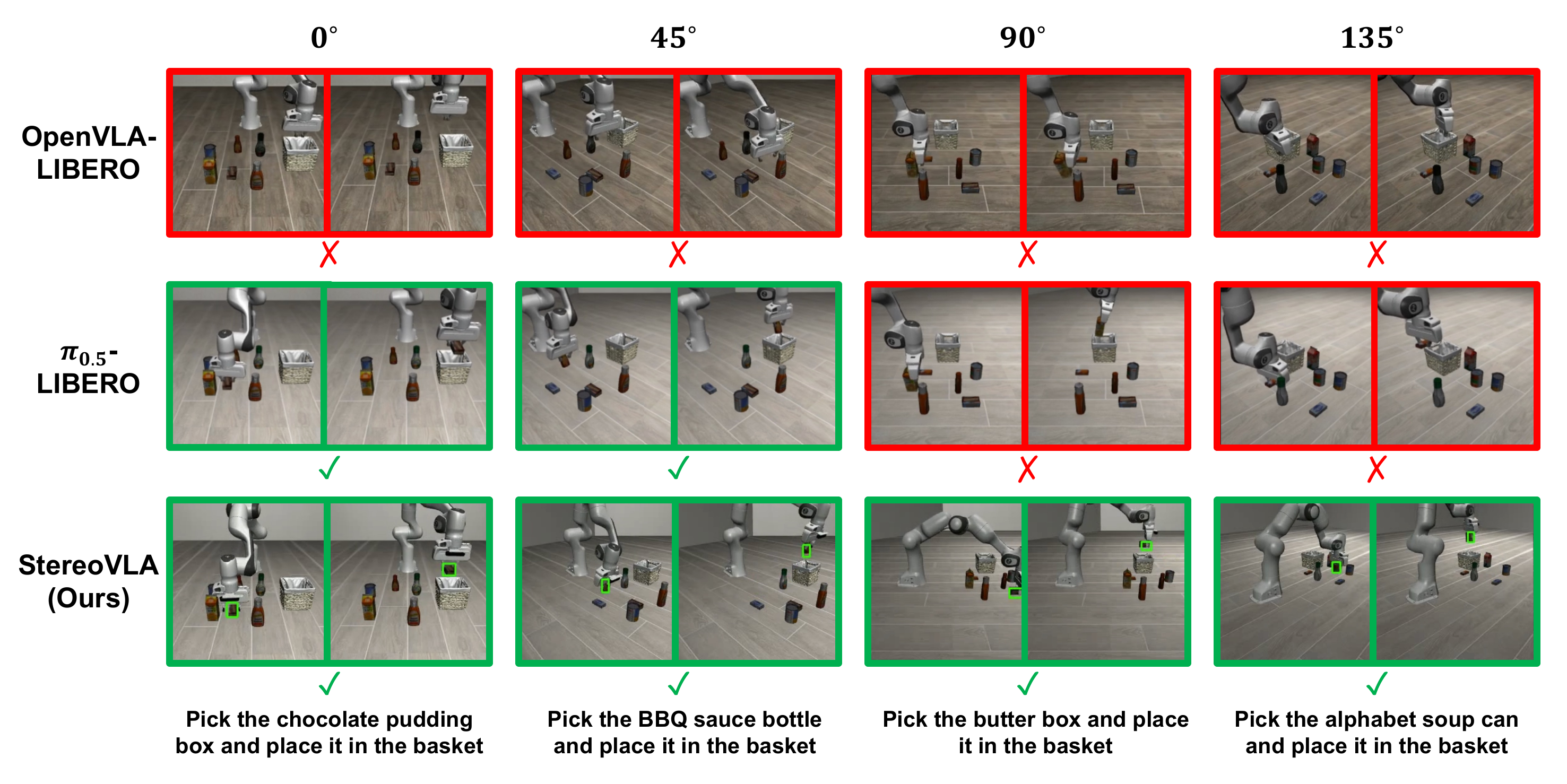}
    \captionof{figure}{
        \textbf{Qualitative comparison on \benchmark.} Each column illustrates a specific manipulation goal in a certain camera viewpoint, while each row corresponds to a distinct model's execution. Successful trials are highlighted with \textcolor{green}{green} borders, whereas failed trials are marked with \textcolor{red}{red} borders. StereoVLA demonstrates superior generalization across all viewpoint angles.
    }
    \label{fig:gallery}
    \vspace{5pt}
\end{strip}

%% file: tables/all_methods.tex
\begin{table}[htbp]
\centering
\caption{\label{table:hyperparams}\textbf{Hyperparameters for all the methods.} SpatialVLA-ED and Spatial-SD share a unified training pipeline, differing only in the use of estimated versus sensor depth during inference. Consequently, they are collectively denoted as *SpatialVLA-D in this table.}
\begin{tabular}{c|c|c}
\toprule
Baseline & Hyperparameter & Value \\ \hline
StereoVLA & batch\_size & 384 \\
     & learning\_rate & 1.6e-4 \\
\midrule
$\pi_{0.5}-S$ & batch\_size & 256 \\
        & learning\_rate & cosine schedule \\
        & warmup\_steps & 1000 \\
        & peak\_lr & 2.5e-5 \\
        & decay\_lr & 2.5e-6 \\
        & decay\_steps & 30000 \\
\midrule
GraspVLA-S & batch\_size & 384 \\
     & learning\_rate & 1.6e-4 \\
\midrule
SpatialVLA   &          batch\_size      &  128     \\
         & learning\_rate & 2e-5 \\
         & image\_aug & true \\
\midrule 
SpatialVLA-*D   &          batch\_size      &  128     \\
         & learning\_rate & 2e-5 \\
         & image\_aug & true \\
\bottomrule
\end{tabular}
\end{table}